\documentclass[sigconf]{acmart}

\renewcommand\footnotetextcopyrightpermission[1]{} 
\usepackage[normalem]{ulem}
\usepackage{amsfonts}       
\usepackage{amsmath}
\usepackage{algorithm}  
\usepackage{algpseudocode}  
\usepackage{multirow}
\usepackage{natbib}
\usepackage{lipsum}
\usepackage{hyperref}

\newcommand\blfootnote[1]{%
  \begingroup
  \renewcommand\thefootnote{}\footnote{#1}%
  \addtocounter{footnote}{-1}%
  \endgroup
}

\begin{document}

\title{ADMM-NN: An Algorithm-Hardware Co-Design Framework of DNNs Using Alternating Direction Method of Multipliers}
\date{}

\author{\normalsize Ao Ren$^{1*}$, ~ Tianyun Zhang$^{2*}$, ~ Shaokai Ye$^{2}$, ~ Jiayu Li$^{2}$, ~ Wenyao Xu$^{3}$, ~ Xuehai Qian$^{4}$, ~ Xue Lin$^{1}$, ~ Yanzhi Wang$^{1}$ \\ 
\normalsize \emph{ $^{*}$ Ao Ren and Tianyun Zhang Contributed equally to this work \\ 
\normalsize $^{1}$ Department of Electrical and Computer Engineering, Northeastern University \\ 
\normalsize $^{2}$ Department of Electrical Engineering and Computer Science, Syracuse University \\
\normalsize $^{3}$ Department of Computer Science and Engineering, SUNY University at Buffalo \\ 
\normalsize $^{4}$ Department of Electrical Engineering, University of Southern California \\
\normalsize $^{1}$ ren.ao@husky.neu.edu, ~ $^{1}\{xue.lin, yanz.wang\}$@northeastern.edu, ~ $^{2}\{tzhan120, sye106, jli221\}$@syr.edu, \\ ~ $^{3}$ wenyaoxu@buffalo.edu, ~ $^{4}$ xuehai.qian@usc.edu \\
}}

\begin{abstract}
\blfootnote{to appear in ASPLOS 2019}
To facilitate efficient embedded and hardware implementations of deep neural networks (DNNs), a number of prior work are dedicated to model compression techniques. The target is to simultaneously reduce the model storage size and accelerate the computation, with minor effect on accuracy. Two important categories of DNN model compression techniques are weight pruning and weight quantization. The former leverages the redundancy in the number of weights, whereas the latter leverages the redundancy in bit representation of weights. These two sources of redundancy can be combined, thereby leading to a higher degree of DNN model compression. However, there lacks a systematic framework of joint weight pruning and quantization of DNNs, thereby limiting the available model compression ratio. Moreover, the computation reduction, energy efficiency improvement, and hardware performance overhead need to be accounted for besides simply model size reduction.

To address these limitations, we present ADMM-NN, the first algorithm-hardware co-optimization framework of DNNs using Alternating Direction Method of Multipliers (ADMM), a powerful technique to deal with non-convex optimization problems with possibly combinatorial constraints. The first part of ADMM-NN is a systematic, joint framework of DNN weight pruning and quantization using ADMM. It can be understood as a smart regularization technique with regularization target dynamically updated in each ADMM iteration, thereby resulting in higher performance in model compression than prior work. The second part is hardware-aware DNN optimizations to facilitate hardware-level implementations. We perform ADMM-based weight pruning and quantization accounting for (i) the computation reduction and energy efficiency improvement, and (ii) the hardware performance overhead due to irregular sparsity. The first requirement prioritizes the convolutional layer compression over fully-connected layers, while the latter requires a concept of the break-even pruning ratio, defined as the minimum pruning ratio of a specific layer that results in no hardware performance degradation. 

Without accuracy loss, we can achieve 85$\times$ and 24$\times$ pruning on LeNet-5 and AlexNet models, respectively, significantly higher than prior work. The improvement becomes more significant when focusing on computation reductions. Combining weight pruning and quantization, we achieve 1,910$\times$ and 231$\times$ reductions in overall model size on these two benchmarks, when focusing on data storage. Highly promising results are also observed on other representative DNNs such as VGGNet and ResNet-50. We release codes and models at anonymous link \textcolor{blue}{http://bit.ly/2M0V7DO}.

\vspace{-3mm}
\end{abstract}
\maketitle

\section{Introduction}

The wide applications of deep neural networks (DNNs), especially for embedded and IoT systems, call for efficient implementations of at least the inference phase of DNNs in power-budgeted systems. To achieve both high performance and energy efficiency, \emph{hardware acceleration of DNNs}, including both FPGA-based and ASIC-based implementations, has been intensively studied both in academia and industry \cite{kwon2018maeri,chen2014diannao,judd2016stripes,sharma2016high,chen2014dadiannao,han2016eie,venkataramani2017scaledeep,umuroglu2017finn,reagen2016minerva,du2015shidiannao,song2018situ,mahajan2016tabla,han2017ese,zhao2017accelerating,suda2016throughput,qiu2016going,chen2017eyeriss,moons201714,desoli201714,whatmough201714,sim201614,bang201714,zhang2016caffeine,zhang2016energy,company1,company2}. With large model size (e.g., for ImageNet dataset \cite{deng2009imagenet}), hardware accelerators suffer from the frequent access to off-chip DRAM due to the limited on-chip SRAM memory. Unfortunately, off-chip DRAM accesses consume significant energy, e.g., 200$\times$ compared to on-chip SRAM \cite{chen2017eyeriss,han2016eie}, and can thus easily dominate the whole system power consumption.

To overcome this hurdle, a number of prior work are dedicated to \emph{model compression} techniques for DNNs, in order to simultaneously reduce the model size (storage requirement) and accelerate the computation, with minor effect on accuracy. Two important categories of DNN model compression techniques are \emph{weight pruning} and \emph{weight quantization}.

A pioneering work of weight pruning is Han \emph{et al.} \cite{han2015learning}, which is an iterative, heuristic method and achieves 9$\times$ reduction in the number of weights in AlexNet (ImageNet dataset). This work has been extended for improving the weight pruning ratio and actual implementation efficiency \cite{han2017ese,guo2017software,han2016eie}. Weight quantization of DNNs has also been investigated in plenty of recent work \cite{leng2017extremely,park2017weighted,zhou2017incremental,lin2016fixed,wu2016quantized,rastegari2016xnor,hubara2016binarized,courbariaux2015binaryconnect,ye2018unified}, quantizing DNN weights to binary values, ternary values, or powers of 2, with acceptable accuracy loss. Both storage and computational requirements are reduced in this way. Multiplication operations may even be eliminated through binary or ternary weight quantizations \cite{rastegari2016xnor,hubara2016binarized,courbariaux2015binaryconnect}.

The effectiveness of weight pruning lies on the redundancy in the number of weights in DNN, whereas the effectiveness of weight quantization is due to the redundancy in bit representation of weights. These two sources of redundancy can be combined, thereby leading to a higher degree of DNN model compression. Despite certain prior work investigating in this aspect using greedy, heuristic method \cite{zhou2017incremental,han2016eie,han2015deep}, there lacks a systematic framework of joint weight pruning and quantization of DNNs. As a result they cannot achieve the highest possible model compression ratio by fully exploiting the degree of redundancy.

Moreover, the prior work on weight pruning and quantization mainly focus on reducing the \emph{model size} of DNNs. As a result, the major model compression is achieved in the fully-connected (FC) layers, which exhibit higher degree of redundancy. On the other hand, the convolutional (CONV) layers, which are the most computationally intensive part of DNNs, do not achieve a significant gain in compression. For example, the pioneering work \cite{han2015learning} achieves only 2.7$\times$ weight reduction in CONV layers for AlexNet model, which still has a high improvement margin when focusing on computation reductions. Furthermore, the weight pruning technique incurs irregularity in weight storage, i.e., the \emph{irregular sparsity}, and corresponding overheads in index storage and calculations, parallelism degradation, etc. These overheads have important impacts in hardware implementations. Take \cite{han2015learning} as an example again. The 2.7$\times$ weight reduction in CONV layers often results in performance degradations as observed in multiple actual hardware implementations \cite{han2017ese,wen2016learning,yu2017scalpel,yang2017designing}.

To address the above limitations, this paper presents ADMM-NN, the first algorithm-hardware co-design framework of DNNs using \emph{Alternating Direction Method of Multipliers (ADMM)}, which is a powerful technique to deal with non-convex optimization problems with possibly combinatorial constraints \cite{ouyang2013stochastic,suzuki2013dual,boyd2011distributed}. The ADMM-NN framework is general, with applications at software-level, FPGA, ASIC, or in combination with new devices and hardware advances.

The first part of ADMM-NN is a systematic, joint framework of DNN weight pruning and quantization using ADMM. Through the application of ADMM, the weight pruning and quantization problems are decomposed into two subproblems: The first is minimizing the loss function of the original DNN with an additional $L_2$ regularization term, and can be solved using standard stochastic gradient descent like ADAM \cite{kingma2016adam}. The second one can be optimally and analytically solved \cite{boyd2011distributed}. The ADMM framework can be understood as a smart regularization technique with regularization target dynamically updated in each ADMM iteration, thereby resulting in high performance in model compression.

The second part of ADMM-NN is hardware-aware optimization of DNNs to facilitate efficient hardware implementations. More specifically, we perform ADMM-based weight pruning and quantization accounting for (i) the computation reduction and energy efficiency improvement, and (ii) the hardware performance overhead due to irregular sparsity. We mainly focus on the model compression on CONV layers, but the FC layers need to be compressed accordingly in order not to cause overfitting (and accuracy degradation). We adopt a concept of the \emph{break-even pruning ratio}, defined as the minimum weight pruning ratio of a specific DNN layer that will not result in hardware performance (speed) degradation. These values are hardware platform-specific. Based on the calculation of such ratios through hardware synthesis (accounting for the hardware performance overheads), we develop efficient DNN model compression algorithm for computation reduction and efficient hardware implementations.

The \textbf{contributions} of this work include: (i) ADMM-based weight pruning, ADMM-based weight quantization solutions of DNNs; (ii) a systematic, joint framework for DNN model compression; and (iii) hardware-aware DNN model compression for computation reduction and efficiency improvement.

Experimental results demonstrate the effectiveness of the proposed ADMM-NN framework. For instance, without any accuracy loss, we can achieve 85$\times$ and 24$\times$ weight pruning on LeNet-5 and AlexNet models, respectively, which are significantly higher than the prior iterative pruning (12$\times$ and 9$\times$, respectively). Combining weight pruning and quantization, we can achieve 1,910$\times$ and 231$\times$ reductions in overall model size on these two benchmarks, when focusing on data storage. Promising results are also observed on other representative DNNs such as VGGNet and ResNet-50. The computation reduction is even more significant compared with prior work. Without any accuracy loss, we can achieve 3.6$\times$ reduction in the amount of computation compared with the prior work \cite{han2015learning,han2015deep}. 
We release codes and models at anonymous link {(\textcolor{blue}{http://bit.ly/2M0V7DO}).

\section{Background}

\subsection{Related Work on Weight Pruning and Quantization}

\begin{figure}
\centering                                    
\includegraphics[width=0.38\textwidth]{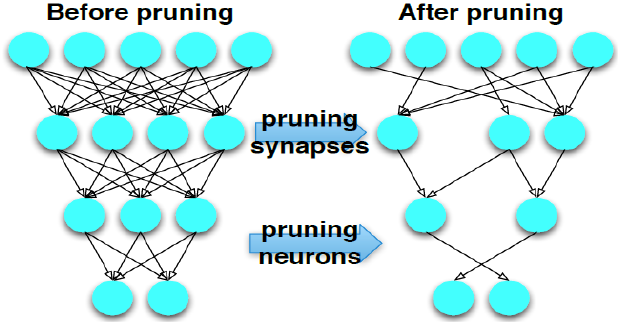} 
\caption{Illustration of weight pruning for DNNs.}\label{Fig:Pruning}
\vspace{-5mm}
\end{figure}

\textbf{Weight pruning} methods leverage the inherent redundancy in the number of weights in DNNs, thereby achieving effective model compression with negligible accuracy loss, as illustrated in Fig. \ref{Fig:Pruning}. A pioneering work of weight pruning is \cite{han2015learning}. It uses a heuristic, iterative method to prune the weights with small magnitudes and retrain the DNN. It achieves 9$\times$ weight reduction on AlexNet for ImageNet dataset, without accuracy degradation. However, this original work achieves relatively low compression ratio (2.7$\times$ for AlexNet) on the CONV layers, which are the key computational part in state-of-the-art DNNs \cite{simonyan2015very,he2016deep}. Besides, indices are needed, at least one per weight, to index the relative location of the next weight. As a result, it suffers from low performance improvement (sometimes even degradation) in actual hardware implementations \cite{wen2016learning,yu2017scalpel,yang2017designing}, when the overhead of irregular sparsity is accounted for.

This work has been extended in two directions.
The first is improving the weight reduction ratio by using more sophisticated heuristics, e.g., incorporating both weight pruning and growing \cite{guo2016dynamic}, using $L_1$ regularization method \cite{wen2016learning}, or genetic algorithm \cite{dai2017nest}. As an example, the recent work NeST \cite{dai2017nest} achieves 15.7$\times$ weight reduction on AlexNet with zero accuracy loss, at the cost of significant training overhead. The second is enhancing the actual implementation efficiency. This goal is achieved by either deriving an effective tradeoff between accuracy and compression ratio, e.g., the \emph{energy-aware pruning} \cite{yang2017designing}, or incorporating regularity and structures into the weight pruning framework, e.g., the \emph{channel pruning} \cite{he2017channel} and \emph{structured sparsity learning} \cite{wen2016learning} approaches.

\textbf{Weight quantization} methods leverage the inherent redundancy in the number of bits for weight representation. Many related work \cite{leng2017extremely,park2017weighted,zhou2017incremental,lin2016fixed,wu2016quantized,rastegari2016xnor,hubara2016binarized,courbariaux2015binaryconnect} present weight quantization techniques to binary values, ternary values, or powers of 2 to facilitate hardware implementations, with acceptable accuracy loss. The state-of-the-art technique adopts an iterative quantization and retraining framework, with randomness incorporated in quantization \cite{courbariaux2015binaryconnect}. It achieves less than 3\% accuracy loss on AlexNet for binary weight quantization \cite{leng2017extremely}. It is also worth noticing that a similar technique, \emph{weight clustering}, groups weights into clusters with arbitrary values. This is different from equal-interval values as in quantization. As a result weight clustering is not as hardware-friendly as quantization \cite{zhu2017trained,han2015deep}.

\textbf{Pros and cons of the two methods:} Weight quantization has clear advantage: it is hardware-friendly. The computation requirement is reduced in proportion to weight representation, and multiplication operations can be eliminated using binary/ternary quantizations. On the other hand, weight pruning incurs inevitable implementation overhead due to the irregular sparsity and indexing \cite{han2015deep,wen2016learning,yu2017scalpel,yang2017designing,ding2017c}.  

The major advantage of weight pruning is the higher potential gain in model compression. The reasons are two folds. First, there is often higher degree of redundancy in the number of weights than bit representation. In fact, reducing each bit in weight presentation doubles the imprecision, which is not the case in pruning. Second, weight pruning performs regularization that strengthens the salient weights and prunes the unimportant ones. It can even increase the accuracy with a moderate pruning ratio \cite{wen2016learning,han2016dsd}. As a result it provides a higher margin of weight reduction. This effect does not exist in weight quantization/clustering. 

\textbf{Combination:} Because they leverage different sources of redundancy, weight pruning and quantization can be effectively combined. However, there lacks a systematic investigation in this direction. The extended work \cite{han2015deep} by Han \emph{et al.} uses a combination of weight pruning and clustering (not quantization) techniques, achieving 27$\times$ model compression on AlexNet. This compression ratio has been updated by the recent work \cite{zhou2017incremental} to 53$\times$ on AlexNet (but without any specification about compressed model).

\subsection{Basics of ADMM}

ADMM has been demonstrated \cite{ouyang2013stochastic,suzuki2013dual} as a powerful tool for solving non-convex optimization problems, potentially with combinatorial constraints.
Consider a non-convex optimization problem that is difficult to solve directly. ADMM method decomposes it into two subproblems that can be solved separately and efficiently. For example, the optimization problem
\begin{equation}
   \min_{\bf{x}}\ \ f({\bf{x}})+g({\bf{x}})
\end{equation}
lends itself to the application of ADMM if $f(\bf{x})$ is differentiable and $g(\bf{x})$ has some structure that can be exploited. Examples of $g(\bf{x})$ include the $L_1$-norm or the indicator function of a constraint set. The problem is first re-written as
\begin{equation}
\label{opt1}
\begin{aligned}
& \min_{\bf{x},\bf{z}}
& & f(\textbf{x})+g(\textbf{z}),
\\ & \text{subject to}
& & \bf{x}=\bf{z}.
\end{aligned}
\end{equation}
Next, by using augmented Lagrangian \cite{boyd2011distributed}, the above problem is decomposed into two subproblems on $\bf{x}$ and $\bf{z}$. The first is $\min_{\bf{x}} f({\bf{x}})+q_1(\bf{x})$, where $q_1(\bf{x})$ is a quadratic function. As $q_1(\bf{x})$ is convex, the complexity of solving subproblem 1 (e.g., via stochastic gradient descent) is the same as minimizing $f(\bf{x})$. Subproblem 2 is $\min_{\bf{z}} g({\bf{z}})+q_2(\bf{z})$, where $q_2(\bf{z})$ is a quadratic function. When function $g$ has some special structure, exploiting the properties of $g$ allows this problem to be solved analytically and optimally. In this way we can get rid of the combinatorial constraints and solve the problem that is difficult to solve directly.

\section{ADMM Framework for Joint Weight Pruning and Quantization}

In this section, we present the novel framework of ADMM-based DNN weight pruning and quantization, as well as the joint model compression problem.

\subsection{Problem Formulation}
Consider a DNN with $N$ layers, which can be convolutional (CONV) and fully-connected (FC) layers. The collection of weights in the $i$-th layer is ${\bf{W}}_{i}$; the collection of bias in the $i$-th layer is denoted by ${\bf{b}}_{i}$. The loss function associated with the DNN is denoted by $f \big( \{{\bf{W}}_{i}\}_{i=1}^N, \{{\bf{b}}_{i} \}_{i=1}^N \big)$.

The problem of weight pruning and quantization is an optimization problem \cite{zhang2018systematic,ye2018unified}:
\begin{equation}
\label{opt0}
\begin{aligned}
& \underset{ \{{\bf{W}}_{i}\},\{{\bf{b}}_{i} \}}{\text{minimize}}
& & f \big( \{{\bf{W}}_{i}\}_{i=1}^N, \{{\bf{b}}_{i} \}_{i=1}^N \big),
\\ & \text{subject to}
& & {\bf{W}}_{i}\in {\bf{S}}_{i}, \; i = 1, \ldots, N.
\end{aligned}
\end{equation}

Thanks to the flexibility in the definition of the constraint set ${\bf{S}}_{i}$, the above formulation is applicable to the individual problems of weight pruning and weight quantization, as well as the joint problem. For the weight pruning problem, the constraint set ${\bf{S}}_{i} =
\left \{ \text{the number of nonzero weights is less than or equal to} ~ \alpha_i \right \}$, where $\alpha_i$ is the desired number of weights after pruning in layer $i$\footnote{An alternative formulation is to use a single $\alpha$ as an overall constraint on the number of weights in the whole DNN.}.
For the weight quantization problem, the set ${\bf{S}}_{i}$=\{the weights in layer i are mapped to the quantization values\} $\left\{Q_1, Q_2, \cdots, Q_M\} \right\}$, where $M$ is the number of quantization values/levels. For quantization, these $Q$ values are fixed, and the interval between two nearest quantization values is the same, in order to facilitate hardware implementations.

For the joint problem, the above two constraints need to be satisfied simultaneously. In other words, the number of nonzero weights should be less than or equal to $\alpha_i$ in each layer, while the remaining nonzero weights should be quantized.

\subsection{ADMM-based Solution Framework}
The above problem is non-convex with combinatorial constraints, and cannot be solved using \emph{stochastic gradient descent} methods (e.g., ADAM \cite{kingma2016adam}) as in original DNN training. But it can be efficiently solved using the ADMM framework (combinatorial constraints can be get rid of.) 
To apply ADMM, we define indicator functions  \begin{eqnarray*}g_{i}({\bf{W}}_{i})=
\begin{cases}
 0 & \text { if } {\bf{W}}_{i}\in {\bf{S}}_{i}, \\ 
 +\infty & \text { otherwise, }
\end{cases}
\end{eqnarray*}
for $i = 1, \ldots, N$.
We then incorporate auxiliary variables ${\bf{Z}}_{i}$ and rewrite problem (\ref{opt0}) as 
\begin{equation}
\label{admm_form}
\begin{aligned}
& \underset{ \{{\bf{W}}_{i}\},\{{\bf{b}}_{i} \}}{\text{minimize}}
& & f \big( \{{\bf{W}}_{i} \}_{i=1}^N, \{{\bf{b}}_{i} \}_{i=1}^N \big)+\sum_{i=1}^{N} g_{i}({\bf{Z}}_{i}),
\\ & \text{subject to}
& & {\bf{W}}_{i}={\bf{Z}}_{i}, \; i = 1, \ldots, N.
\end{aligned}
\end{equation}

Through application of the augmented Lagrangian \cite{boyd2011distributed}, problem (\ref{admm_form}) is decomposed into two subproblems by ADMM. We solve the subproblems iteratively until convergence. The first subproblem is
\begin{equation}
\label{4}
 \underset{ \{{\bf{W}}_{i}\},\{{\bf{b}}_{i} \}}{\text{minimize}}
\ \ \ f \big( \{{\bf{W}}_{i} \}_{i=1}^N, \{{\bf{b}}_{i} \}_{i=1}^N \big)+\sum_{i=1}^{N} \frac{\rho_{i}}{2}  \| {\bf{W}}_{i}-{\bf{Z}}_{i}^{k}+{\bf{U}}_{i}^{k} \|_{F}^{2}, \\
\end{equation}
where ${\bf{U}}_{i}^{k}$ is the dual variable updated in each iteration, ${\bf{U}}_{i}^{k}:={\bf{U}}_{i}^{k-1}+{\bf{W}}_{i}^{k}-{\bf{Z}}_{i}^{k}$. In the objective function of (\ref{4}), the first term is the differentiable loss function of DNN, and the second quadratic term is differentiable and convex. The combinatorial constraints are effectively get rid of. This problem can be solved by stochastic gradient descent (e.g., ADAM) and the complexity is the same as training the original DNN.

The second subproblem is
\begin{equation}
\label{5}
 \underset{ \{{\bf{Z}}_{i} \}}{\text{minimize}}
\ \ \ \sum_{i=1}^{N} g_{i}({\bf{Z}}_{i})+\sum_{i=1}^{N} \frac{\rho_{i}}{2} \| {\bf{W}}_{i}^{k+1}-{\bf{Z}}_{i}+{\bf{U}}_{i}^{k} \|_{F}^{2}. \\
\end{equation}
As $g_{i}(\cdot)$ is the indicator function of ${\bf{S}}_{i}$, the analytical solution of subproblem (\ref{5}) is 
\begin{equation}
\label{6}
  {\bf{Z}}_{i}^{k+1} = {{\Pi}_{{\bf{S}}_{i}}}({\bf{W}}_{i}^{k+1}+{\bf{U}}_{i}^{k}),
\end{equation}
where ${{\Pi}_{{\bf{S}}_{i}}(\cdot)}$ is Euclidean projection of ${\bf{W}}_{i}^{k+1}+{\bf{U}}_{i}^{k}$ onto the set ${\bf{S}}_{i}$. The details of the solution to this subproblem is problem-specific. For weight pruning and quantization problems, the optimal, analytical solutions of this problem can be found. The derived ${\bf{Z}}_{i}^{k+1}$ will be fed into the first subproblem in the next iteration.

The intuition of ADMM is as follows. In the context of DNNs, the ADMM-based framework can be understood as a smart regularization technique. Subproblem 1 (Eqn. (\ref{4})) performs DNN training with an additional $L_2$ regularization term, and the regularization target ${\bf{Z}}_{i}^{k}-{\bf{U}}_{i}^{k}$ is dynamically updated in each iteration through solving subproblem 2. This dynamic updating process is the key reason why ADMM-based framework outperforms conventional regularization method in DNN weight pruning and quantization.

\subsection{Solution to Weight Pruning and Quantization, and the Joint Problem}

Both weight pruning and quantization problems can be effectively solved using the ADMM framework. For the weight pruning problem, the Euclidean projection Eqn. (\ref{6}) is to keep $\alpha_i$ elements in ${\bf{W}}_{i}^{k+1}+{\bf{U}}_{i}^{k}$ with largest magnitude and set the rest to be zero \cite{ouyang2013stochastic,suzuki2013dual}. This is proved to be the optimal and analytical solution to subproblem 2 (Eqn. (\ref{5})) in weight pruning. 

For the weight quantization problem, the Euclidean projection Eqn. (\ref{6}) is to set every element in ${\bf{W}}_{i}^{k+1}+{\bf{U}}_{i}^{k}$ to be the quantization value closest to that element. This is also the optimal and analytical solution to subproblem 2 in quantization. The determination of quantization values will be discussed in details in the next subsection.

For both weight pruning and quantization problems, the first subproblem has the same form when ${\bf{Z}}_{i}^{k}$ is determined through Euclidean projection. As a result they can be solved in the same way by stochastic gradient descent (e.g., the ADAM algorithm).

\begin{figure}
\centering                                    
\includegraphics[width=0.4\textwidth]{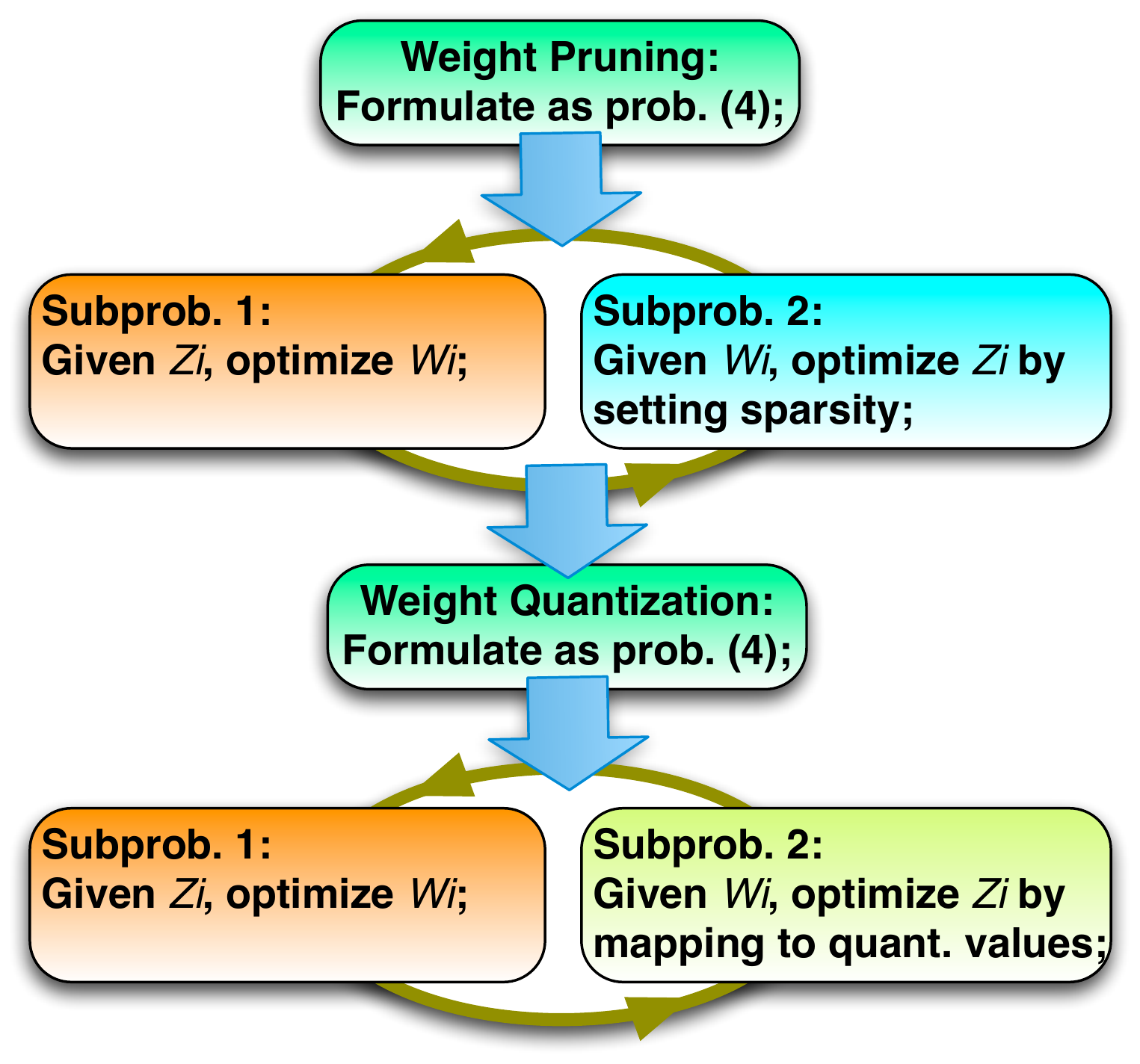} 
\caption{Algorithm of joint weight pruning 
and quantization using ADMM.}\label{Fig:AlgorithmOverview}
\vspace{-4mm}
\end{figure}

For the joint problem of weight pruning and quantization, there is an additional degree of flexibility when performing Euclidean projection, i.e., a specific weight can be either projected to zero or to a closest quantization value. This flexibility will add difficulty in optimization. To overcome this hurdle, we perform weight pruning and quantization in two steps. We choose to perform weight pruning first, and then implement weight quantization on the remaining, non-zero weights. The reason for this order is the following observation: There typically exists higher degree of redundancy in the number of weights than the bit representation of weights. As a result, we can typically achieve higher model compression degree using weight pruning, without any accuracy loss, compared with quantization. The observation is validated by prior work \cite{han2017ese,guo2017software,han2016eie} (although many are on clustering instead of quantization), and in our own investigations. Fig. \ref{Fig:AlgorithmOverview} summaries the key steps of solving the joint weight pruning and quantization problem based on ADMM framework.

Thanks to the fast theoretical convergence rate of ADMM, the proposed algorithms have fast convergence. To achieve a good enough compression ratio for AlexNet, we need 72 hours for weight pruning and 24 hours for quantization. This is much faster than \cite{han2015learning} that requires 173 hours for weight pruning only.

\subsection{Details in Parameter Determination}

\subsubsection{Determination of Weight Numbers in Pruning:}

The most important parameters in the ADMM-based weight pruning step are the $\alpha_i$ values for each layer $i$. To determine these values, we start from the values derived from the prior weight pruning work \cite{han2015learning,han2015deep}. When targeting high compression ratio, we reduce the $\alpha_i$ values proportionally for each layer. When targeting computation reductions, we deduct the $\alpha_i$ values for convolutional (CONV) layers, because CONV layers account for the major computation compared with FC layers. Our experimental results demonstrate about 2-3$\times$ further compression under the same accuracy, compared with the prior work \cite{han2015learning,han2015deep,yu2017compressing,dong2017learning}.

The additional parameters in ADMM-based weight pruning, i.e., the penalty parameters $\rho_i$, are set to be $\rho_{1}=\dots=\rho_{N} = 3\times 10^{-3}$. This choice is basically very close for different DNN models, such as AlexNet \cite{krizhevsky2012imagenet} and VGG-16 \cite{simonyan2014very}. The pruning results are not sensitive to the penalty parameters of the optimal choice, unless these parameters are increased or decreased by orders of magnitude.

\begin{table*}[h]
\centering
\caption{Weight pruning ratio and accuracy on the LeNet-5 model for MNIST dataset by our ADMM-based framework and other benchmarks.}\label{Table:LeNet}
\begin{tabular}{p{4cm}p{3.cm}p{3.7cm}p{3.6cm}}
\hline
Benchmark &  Top 1 accuracy  & Number of parameters  & Weight pruning ratio  \\ 
\hline
Original LeNet-5 Model & 99.2\% &  430.5K   & 1$\times$ \\
\hline
\bf{Our Method} & 99.2\% & 5.06K  & 85$\times$ \\
\hline
\bf{Our Method} & 99.0\% & 2.58K  & 167$\times$ \\
\hline
Iterative pruning \cite{han2015learning} & 99.2\% & 35.8K  &  12$\times$\\
\hline
Learning to share \cite{zhang2018learning} & 98.1\% & 17.8K  &  24.1$\times$ \\
\hline
Net-Trim \cite{aghasi2017net} & 98.7\% & 9.4K & 45.7$\times$\\
\hline
\end{tabular}
\end{table*}

\begin{table*}[h]
\centering
\caption{Weight pruning ratio and accuracy on the AlexNet model for ImageNet dataset by our ADMM-based framework and other benchmarks.}\label{Table:AlexNet}
\begin{tabular}{p{4cm}p{2.5cm}p{2.5cm}p{3.3cm}p{3.3cm}}
\hline
Benchmark &  Top 1 accuracy  & Top 5 accuracy & Number of parameters  & Weight pruning ratio  \\ 
\hline
Original AlexNet Model & 57.2\% & 80.2\% &  60.9M   & 1$\times$ \\
\hline
\bf{Our Method} & 57.1\% & 80.2\% & 2.5M & 24$\times$ \\
\hline
\bf{Our Method} & 56.8\% & 80.1\% & 2.05M & 30$\times$ \\
\hline
Iterative pruning \cite{han2015learning} & 57.2\% & 80.3\% & 6.7M  &  9$\times$\\
\hline
Low rank \& sparse \cite{yu2017compressing} & 57.3\% & 80.3\% & 6.1M & 10$\times$\\
\hline
Optimal Brain Surgeon \cite{dong2017learning} & 56.9\% & 80.0\% & 6.7M & 9.1$\times$\\
\hline
SVD \cite{denton2014exploiting} & - & 79.4\% & 11.9M & 5.1$\times$\\
\hline
NeST \cite{dai2017nest} & 57.2\% & 80.3\% & 3.9M & 15.7$\times$\\
\hline
\end{tabular}
\end{table*}

\begin{table*}[h]
\centering
\caption{Weight pruning ratio and accuracy on the VGGNet model for ImageNet dataset by our ADMM-based framework and other benchmarks.}\label{Table:Vgg}
\begin{tabular}{p{4cm}p{2.5cm}p{2.5cm}p{3.3cm}p{3.3cm}}
\hline
Benchmark &  Top 1 accuracy  & Top 5 accuracy & Number of parameters  & Weight pruning ratio  \\ 
\hline
Original VGGNet Model & 69.0\% & 89.1\% &  138M   & 1$\times$ \\
\hline
\bf{Our Method} & 68.7\% & 88.9\% & 5.3M & 26$\times$ \\
\hline
\bf{Our Method} & 69.0\% & 89.1\% & 6.9M & 20$\times$ \\
\hline
Iterative pruning \cite{han2015learning} & 68.6\% & 89.1\% & 10.3M  &  13$\times$\\
\hline
Low rank \& sparse \cite{yu2017compressing} & 68.8\% & 89.0\% & 9.2M & 15$\times$\\
\hline
Optimal Brain Surgeon \cite{dong2017learning} & 68.0\% & 89.0\% & 10.4M & 13.3$\times$\\
\hline
\end{tabular}
\end{table*}

\begin{table*}[h]
\centering
\caption{Weight pruning ratio and accuracy on the ResNet-50 model for ImageNet dataset.}\label{Table:ResNet}
\begin{tabular}{p{4cm}p{3.3cm}p{3.3cm}p{3.3cm}}
\hline
Benchmark & Accuracy degradation & Number of parameters  & Weight pruning ratio  \\ 
\hline
Original ResNet-50 Model & 0.0\% &  25.6M   & 1$\times$ \\
\hline
Fine-grained Pruning \cite{mao2017exploring} & 0.0\% & 9.8M & 2.6$\times$ \\
\hline
\bf{Our Method} & 0.0\% & 3.6M & 7$\times$ \\
\hline
\bf{Our Method} & 0.3\% & 2.8M & 9.2$\times$ \\
\hline
\bf{Our Method} & 0.8\% & 1.47M & 17.4$\times$ \\
\hline
\end{tabular}
\end{table*}

\subsubsection{Determination of Quantization Values:}

After weight pruning is performed, the next step is weight quantization on the remaining, non-zero weights. We use $n$ bits for equal-distance quantization to facilitate hardware implementations, which means there are a total of $M=2^n$ quantization levels. More specifically, for each layer $i$, we quantize the weights into a set of quantization values $\displaystyle\{-\frac{M}{2}q_i, ..., -2 q_i, -q_i, q_i, 2q_i, ..., \frac{M}{2}q_i    \}$. 
Please note that 0 is not a quantization value because it means that the corresponding weight has been pruned.

\begin{figure}
\centering                                    
\includegraphics[width=0.48\textwidth]{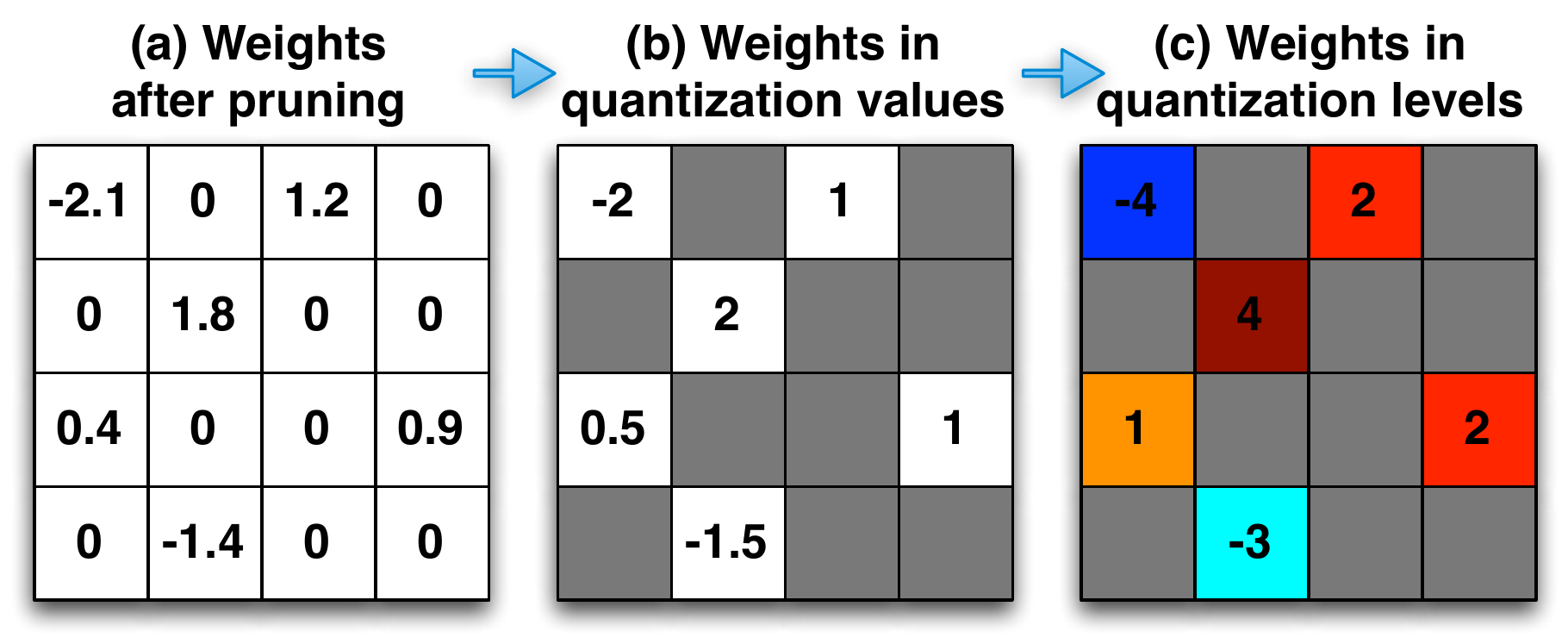} 
\caption{Illustration of weight quantization (the interval value $q_i=0.5$).}\label{Fig:Quant}
\vspace{-5mm}
\end{figure}

The \emph{interval} $q_i$ is the distance between two adjacent quantization values, and may be different for different layers. This is compatible with hardware implementations. This is because (i) the $q_i$ value of each layer is stored along with the quantized weights of that specific layer, and (ii) a scaling computation will be performed using the $q_i$ value on the outputs of layer $i$. Such scaling computation is needed in equal-distance weight quantization \cite{lin2016fixed,wu2016quantized}.

Fig. \ref{Fig:Quant} shows an example of weight quantization processure. Suppose we have a $4\times 4$ weight matrix. Fig. \ref{Fig:Quant} (a) is the weights to be quantized, obtained after pruning. Based on the weight values, $q_i = 0.5$, $n=3$, and $M=2^n$ are determined. Fig. \ref{Fig:Quant} (b) is the weight values after quantization, and Fig. \ref{Fig:Quant} (c) is the weights represented in quantization levels. Note that quantization levels encoded in binary bits are the operands to be stored and operated in the hardware. For the case of Fig. \ref{Fig:Quant}, quantization levels $\{-4,-3,-2,-1,1,2,3,4\}$ are encoded using 3 binary bits, since $0$ denoting pruned weights is not needed. Weights in quantization levels (Fig. \ref{Fig:Quant} (c) ) times $q_i=0.5$ resulting in quantized weights (Fig. \ref{Fig:Quant} (b) ).

The interval $q_i$ and number of quantization levels $M$ ($n$) are pre-defined, and should be determined in an effective manner. For $M$ ($n$) values, we start from the results of some prior work like \cite{han2015learning}, and reduce $n$ accordingly. For example, \cite{han2015deep} uses on average around 5 bits for quantization (essentially clustering) in AlexNet, whereas our results show that 3-4 bits on average are sufficient in quatization without incurring any accuracy loss, on representative benchmark DNNs.

To determine $q_i$, let $w_i^j$ denote the $j$-th weight in layer $i$, and $f(w_i^j)$ denote the quantization function to the closest quantization value. Then the total square error in a single quantization step is given by $\sum_j\big|w_i^j-f(w_i^j)\big|^2$. We derive $q_i$ using binary search method, such that the above total square error is minimized. In this way we determine both $q_i$ and $M$ ($n$) for weight quantization.

\section{Results and Discussions on DNN Model Compressions}

In this section, we summarize the software-level results on DNN model compression using the proposed ADMM framework of weight pruning and quantization. We perform testing on a set of representative DNN benchmarks, LeNet-5 \cite{lecun2015lenet} for MNIST dataset, AlexNet \cite{krizhevsky2012imagenet} (BVLC model and CaffeNet model, both open-source), VGGNet \cite{simonyan2014very}, and ResNet-50 \cite{he2016deep} for ImageNet dataset. We initialize ADMM using pretrained DNN models and then perform weight pruning/quantization. We focus on the model compression of \underline{the overall DNN model} (i.e., the total number of weights and total number of bits for weight representations). We perform comparison with representative works on DNN weight pruning and quantization (clustering), and demonstrate the significant improvement using the proposed ADMM framework. Algorithm implementations are on the open-source Caffe tool with code/model release, and DNN training and compression are performed using NVIDIA Tesla P100 and GeForce GTX 1080Ti GPUs.

\subsection{Results on ADMM-based Weight Pruning}

Table \ref{Table:LeNet} shows the weight pruning results on the LeNet-5 model, in comparison with various benchmarks. LeNet-5 contains two CONV layers, two pooling layers, and two FC layers, and can achieve 99.2\% test accuracy on the MNIST dataset. Our ADMM-based weight pruning framework does not incur accuracy loss and can achieve a much higher weight pruning ratio on these networks compared with the prior iterative pruning heuristic \cite{han2015learning}, which reduces the number of weights by 12$\times$ on LeNet-5. In fact, our pruning method reduces the number of weights by 85$\times$, which is 7.1$\times$ improvement compared with \cite{han2015learning}. The maximum weight reduction is 167$\times$ for LeNet-5 when the accuracy is as high as 99.0\%.

Similar results can be achieved on the BVLC AlexNet model and VGGNet model on the ImageNet ILSVRC-2012 dataset. The original BVLC AlexNet model can achieve a top-1 accuracy 57.2\% and a top-5 accuracy 80.2\% on the validation set, containing 5 CONV (and pooling) layers and 3 FC layers with a total of 60.9M parameters. The original VGGNet model achieves a top-1 accuracy 69.0\% and top-5 accuracy 89.1\% on ImageNet dataset, with a total of 138M parameters. Table \ref{Table:AlexNet} shows the weight pruning comparison results on AlexNet while Table \ref{Table:Vgg} shows the comparison results on VGGNet. The proposed ADMM method can achieve 24$\times$ weight reduction in AlexNet and 26$\times$ weight reduction in VGGNet, without any accuracy loss. These results are at least twice as the state-of-the-art, and clearly demonstrate the advantage of the proposed weight pruning method using ADMM.

For the results on ResNet-50 model on ImageNet as shown in Table \ref{Table:ResNet}, we achieve 7$\times$ weigh pruning without accuracy degradation, and 17.4$\times$ with minor accuracy degradation less than 1\%.

The reasons for the advantage are two folds: First, the ADMM-based framework is a systematic weight pruning framework based on optimization theory, which takes an overall consideration of the whole DNN instead of making local, greedy pruning choices. In fact, with a moderate pruning ratio of 3$\times$, the top-1 accuracy of AlexNet can be even increased to 59.1\%, almost 2\% increase. Second, as discussed before, the ADMM-based framework can be perceived as a smart, dynamic DNN regularization technique, in which the regularization target is analytically adjusted in each iteration. This is very different from the prior regularization techniques \cite{wen2016learning,goodfellow2016deep} in which the regularization target is predetermined and fixed.

\subsection{Results on ADMM-based Joint Weight Pruning and Quantization for DNNs}

\begin{table*}[h]
\centering
\caption{Model size compression ratio on the LeNet-5 model for MNIST dataset by our ADMM-based framework and baseline.}\label{Table:LeNetwQuantization}
\begin{tabular}{p{3.cm}p{1.5cm}p{1.5cm}p{1.5cm}p{1.2cm}p{3cm}p{3cm}}
\hline
Benchmark &  Accuracy degrade  & Para. No.  & CONV quant. & FC quant. & Total data size/ \quad Compress ratio & Total model size \quad (including index)/ Compress ratio \\ 
\hline
Original LeNet-5 & 0.0\% &  430.5K   & 32b & 32b & 1.7MB  & 1.7MB \\
\hline
\bf{Our Method} & 0.2\% & 2.57K  & 3b & 2b & 0.89KB / 1,910$\times$ & 2.73KB / 623$\times$ \\
\hline
Iterative pruning \cite{han2015deep} & 0.1\% & 35.8K  & 8b & 5b & 24.2KB / 70.2$\times$ & 52.1KB / 33$\times$\\
\hline
\end{tabular}
\end{table*}

\begin{table*}[h]
\centering
\caption{Model size compression ratio on the AlexNet, VGGNet, and ResNet-50 models for ImageNet dataset by our ADMM-based framework and baselines.}\label{Table:AlexNetVGGwQuantization}
\begin{tabular}{p{3.3cm}p{1.5cm}p{1.5cm}p{1.5cm}p{1.2cm}p{3cm}p{3cm}}
\hline
Benchmark &  Accuracy degrade  & Para. No.  & CONV quant. & FC quant. & Total data size/ \quad Compress ratio & Total model size \quad (including index)/ Compress ratio \\ 
\hline
Original AlexNet & 0.0\% &  60.9M   & 32b & 32b & 243.6MB  & 243.6MB \\
\hline
\bf{Our Method} & 0.2\% & 2.25M  & 5b & 3b & 1.06MB / 231 $\times$ & 2.45MB / 99$\times$ \\
\hline
Iterative pruning \cite{han2015deep} & 0.0\% & 6.7M  & 8b & 5b & 5.4MB / 45$\times$ & 9.0MB / 27$\times$\\
\hline
Binary quant. \cite{leng2017extremely} & 3.0\% & 60.9M  & 1b & 1b & 7.3MB / 32$\times$ & 7.3MB / 32$\times$\\
\hline
Ternary quant. \cite{leng2017extremely} & 1.8\% & 60.9M  & 2b & 2b & 15.2MB / 16$\times$ & 15.2MB / 16$\times$\\
\hline
\hline
Original VGGNet & 0.0\% &  138M   & 32b & 32b & 552MB  & 552MB \\
\hline
\bf{Our Method} & 0.1\% & 6.9M  & 5b & 3b & 3.2MB / 173$\times$ & 8.3MB / 66.5$\times$ \\
\hline
Iterative pruning \cite{han2015deep} & 0.0\% & 10.3M  & 8b & 5b & 8.2MB / 67$\times$ & 17.8MB / 31$\times$\\
\hline
Binary quant. \cite{leng2017extremely} & 2.2\% & 138M  & 1b & 1b & 17.3MB / 32$\times$ & 17.3MB / 32$\times$\\
\hline
Ternary quant. \cite{leng2017extremely} & 1.1\% & 138M  & 2b & 2b & 34.5MB / 16$\times$ & 34.5MB / 16$\times$\\
\hline
\hline
Original ResNet-50 & 0.0\% &  25.6M   & 32b & 32b & 102.4MB  & 102.4MB \\
\hline
\bf{Our Method} & 0.0\% & 3.6M  & 6b & 6b & 2.7MB / 38$\times$ & 4.1MB / 25.3$\times$ \\
\hline
\bf{Our Method} & 2.0\% & 1.47M  & 4b & 4b & 0.73MB / 140$\times$ & 1.65MB / 62$\times$\\
\hline
\end{tabular}
\end{table*}

In this section we perform comparisons on the joint weight pruning and quantization results. Table \ref{Table:LeNetwQuantization} presents the results on LeNet-5, while Table \ref{Table:AlexNetVGGwQuantization} presents the results on AlexNet, VGGNet, and ResNet-50. We can simultaneously achieve 167$\times$ pruning ratio on LeNet-5, with an average of 2.78-bit for weight representation (fewer-bit representation for FC layers and more-bit for CONV layers). When accounting for the weight data representation only, the overall compression ratio is 1,910$\times$ on LeNet-5 when comparing with 32-bit floating point representations. For weight data representation, only 0.89KB is needed for the whole LeNet-5 model with 99\% accuracy. This is clearly approaching the theoretical limit considering the input size of 784 pixels (less than 1K) for each MNIST data sample.

For AlexNet and VGGNet models, we can use an average of 3.7-bit for weight representation. When accounting for the weight data only, the overall compression ratios are close to 200$\times$. These results are significantly higher than the prior work such as \cite{han2015learning,han2015deep}, even when \cite{han2015deep} focuses on weight clustering instead of quantization\footnote{Weight clustering is less hardware-friendly, but should perform better than weight quantization in model compression. The reason is because weight quantization can be perceived as a special case of clustering.}. For example, \cite{han2015learning} achieves 9$\times$ weight pruning on AlexNet and uses an average of higher than 5 bits (8 bits for CONV layers and 5 bits for FC layers) for weight representation. These results are also higher than performing weight quantization/clustering alone because the maximum possible gain when performing quantization/clustering alone is 32 (we need to use 1 bit per weight anyway) compared with floating-point representations, let alone accuracy degradations. These results clearly demonstrate the effectiveness of the proposed ADMM framework on joint weight pruning and quantization for DNNs. Similar results are also observed on the joint weight pruning and quantization results on ResNet-50 model.

However, we must emphasize that the actual storage reduction cannot reach such a high gain. For DNNs, the \emph{model size} is defined as the total number of bits (or Bytes) to actually store a DNN model. The reason for this gap is the indices, which are needed (at least) one per weight with weight pruning in order to locate the ID of the next weight \cite{han2015learning}. For instance, we need more bits for each index for the pruned AlexNet than \cite{han2015learning} because we achieve a higher pruning ratio. The storage requirement for indices will be even higher compared with the actual data, because the ADMM framework is very powerful in weight quantization. This will add certain overhead for the overall model storage, as also shown in the tables.

\begin{table}[h]
\centering
\caption{Layer-wise weight pruning results on the AlexNet model without accuracy loss using the ADMM framework.}\label{Table:AlexNetLayer}
\begin{tabular}{p{1.5cm}p{1.0cm}p{1.5cm}p{2.5cm}}
\hline
Layer &  Para. No.  & Para. No. after prune  & Para. Percentage after prune  \\ 
\hline
conv1 & 34.8K & 28.19K & 81\% \\
conv2 & 307.2K & 61.44K & 20\% \\
conv3 & 884.7K & 168.09K & 19\% \\
conv4 & 663.5K & 132.7K & 20\% \\
conv5 & 442.4K & 88.48K & 20\% \\
fc1   & 37.7M  & 1.06M & 2.8\% \\
fc2   & 16.8M  & 0.99M & 5.9\% \\
fc3   &4.1M  & 0.38M & 9.3\% \\
\hline
total &60.9M &2.9M & 4.76\% \\
\hline
\end{tabular}
\vspace{-2mm}
\end{table}

Finally, we point out that it may be somewhat biased when only considering the model size reduction of DNNs. We list in Table \ref{Table:AlexNetLayer} the layer-wise weight pruning results for AlexNet, using the proposed ADMM framework. We can observe that the major weight pruning and quantization are achieved in the FC layers, compared with CONV layers. The reasons are that the FC layers account for more than 90\% of weights and possess a higher degree of redundancy, thereby enabling higher degree of weight pruning/quantization. This is the same as the prior work such as \cite{han2015learning}, which achieves 9$\times$ overall weight reduction while only 2.7$\times$ reduction on CONV layers. It uses 5-bit for weight representation of FC layers and 8 bits for CONV layers. On the other hand, we emphasize that the CONV layers account for the major computation in state-of-the-art DNNs, e.g., 95\% to 98\% in AlexNet and VGGNet \cite{krizhevsky2012imagenet,simonyan2014very}, and even more for ResNet \cite{he2016deep}. For computation reduction and energy efficiency improvement, it is more desirable to focus on CONV layers for weight pruning and quantization. This aspect will be addressed in the next section.

\subsection{Making AlexNet and VGGNet On-Chip}

An important indication of the proposed ADMM framework is that the weights of most of the large-scale DNNs can be stored on-chip for FPGA and ASIC designs. Let us consider AlexNet and VGGNet as examples. For AlexNet, the number of weights before pruning is 60.9M, corresponding to 244MB storage (model size) when 32-bit floating point number is utilized for weight representation. Using the proposed ADMM weight pruning and quantization framework, the total storage (model size) of AlexNet is reduced to 2.45MB (using 2.25M weights) when the indices are accounted for. This model size is easily accommodated by the medium-to-high end FPGAs, such as Xilinx Kintex-7 series, and ASIC designs. This is achieved without any accuracy loss.

On the other hand, VGGNet, as one of the largest DNNs that is widely utilized, has a total number of 138M weights, corresponding to 552MB storage when 32-bit floating point number is used for weight representation. Using the proposed ADMM framework, the total model size of VGGNet is reduced to 8.3MB (using 6.9M weights) when the indices are accounted for. This model size can still be accommodated by a single high-end FPGA such as Altera (Intel) DE-5 and Xilinx Virtex-7.
The effect that large-scale AlexNet and VGGNet models can be stored using on-chip memory of single FPGA/ASIC will significantly facilitate the wide application of large-scale DNNs, in embedded, mobile, and IoT systems. It can be a potential game changer. On the other hand, when accounting for the computation reductions rather than mere storage (model size) reduction, it is more desirable to focus mainly on the model compression on CONV layers rather than the whole DNN model. Also it is desirable to focus more on CONV layers since a smaller on-chip memory can be both cost and speed-beneficial, which is critical especially for custom ASIC.

\section{Hardware-Aware Computation Reduction}


\textbf{Motivation} 
As discussed in the previous section and illustrated in Table \ref{Table:AlexNetLayer}, the current gains in weight pruning and quantization are mainly attributed to the redundancy in FC layers. This optimization target is not the most desirable when accounting for the computation reduction and energy efficiency improvement. The reason is that CONV layers account for the major computation in state-of-the-art DNNs, even reaching 98\% to 99\% for the recent VGGNet and ResNet models \cite{krizhevsky2012imagenet,simonyan2014very}. In actual ASIC design and implementations, it will be desirable to allocate on-chip memory for the compressed CONV layers while using off-chip memory for the less computationally intensive FC layers. In this way the on-chip memory can be reduced, the memory speed can be faster, while the major computation part of DNN (CONV layers) can be accelerated. Therefore it is suggested to perform weight pruning and quantization focusing on the CONV layers.  

The prior weight pruning work \cite{han2015learning,han2015deep} cannot achieve a satisfactory weight pruning ratio on CONV layers while guaranteeing the overall accuracy. For example, \cite{han2015learning} achieves only 2.7$\times$ weight pruning on the CONV layers of AlexNet. In fact, the highest gain in reference work on CONV layer pruning is 5.0$\times$ using $L_1$ regularization \cite{wen2016learning}, and does not perform any pruning on FC layers. Sometimes, a low weight pruning ratio will result in hardware performance degradation, as reported in a number of actual hardware implementations \cite{wen2016learning,yu2017scalpel,yang2017designing}. The key reason is the irregularity in weight storage, the associated overhead in calculating weight indices, and the degradation in parallelism. This overhead is encountered in the PE (processing element) design when sparsity (weight pruning) is utilized. This performance overhead needs to be accurately characterized and effectively accounted for in the hardware-aware weight pruning framework.
\vspace{-2.5mm}

\subsection{Algorithm-Hardware Co-Optimization}

In a nutshell, we need to (i) focus mainly on CONV layers in weight pruning/quantization, and (ii) effectively account for the hardware performance overhead for irregular weight storage, in order to facilitate efficient hardware implementations. We start from an observation about coordinating weight pruning in CONV and FC layers for maintaining overall accuracy.

\textbf{\emph{Observation on Coordinating Weight Pruning:}} Even when we focus on CONV layer weight pruning, we still need to prune the FC layers moderately (e.g., about 3-4$\times$) for maintaining the overall accuracy. Otherwise it will incur certain accuracy degradation.

Although lack of formal proof, the observation can be intuitively understood in the following way: The original DNN models, such as LeNet-5, AlexNet, or VGGNet, are heavily optimized and the structures of CONV and FC layers match each other. Pruning the CONV layers alone will incur mismatch in structure and number of weights with the FC layers, thereby incurring overfitting and accuracy degradation. This is partially the reason why prior work like $L_1$ regularization \cite{wen2016learning} does not have satisfactory performance even when only focusing on CONV layers. This observation brings 0.5\% to 1\% accuracy improvement, along with additional benefit of simultaneous computation reduction and model size reduction, and will be exploited in our framework.

\textbf{\emph{Break-even Weight Pruning Ratio:}} Next, we define the concept of \emph{break-even weight pruning ratio}, as the minimum weight pruning ratio of a specific (CONV or FC) layer that will not result in hardware performance degradation. Below this break-even ratio, performance degradation will be incurred, as actually observed in \cite{wen2016learning,yu2017scalpel,yang2017designing}. This break-even pruning ratio is greater than 1 because of the hardware performance overhead from irregular sparsity. It is hardware platform-specific. It is important to the hardware-aware weight pruning framework. For example, if the actual weight pruning ratio for a specific layer is lower than the break-even ratio, there is no need to perform weight pruning on this layer. In this case, we will restore the original structure of this layer and this will leave more margin for weight pruning in the other layers with more benefits.

\textbf{\emph{Break-even Pruning Ratio Calculation:}} To calculate the break-even pruning ratios, we fairly compare (i) the inference delay of the hardware implementation of the original DNN layer without pruning with (ii) the delays of hardware implementations under various pruning ratios. The comparison is under the same hardware area/resource. We control two variables: (i) a predefined, limited hardware area, and (ii) the goal to complete all computations in one DNN layer, which will be different under various pruning ratios. Specifically, we set the hardware implementation of the original layer as baseline, thus its hardware area becomes a hard limit. Any hardware implementations supporting weight pruning cannot exceed this limit.

Hardware resources of the baseline consist of two parts: one is process elements (PE) responsible for GEMM (general matrix multiplication) and activation calculations, and the other is SRAM that stores features, weights, and biases. Although the implementations under various pruning ratios are also composed of PEs and SRAMs, the differences lie in three aspects: (i) the area occupied by SRAM is different. This is because with different pruning ratios, the numbers of indices are different, and the numbers of weights are different as well; (ii) the remaining resources for PE implementation are thus different. It is possible to have more resources for PE implementation or less; (iii) the maximum frequency of each type of implementations is different, due to the difference in the size of PEs and index decoding components. 

Being aware of these differences, we implement the baseline and 9 pruning cases with pruning portions ranging from 10\% to 90\%. We adopt the state-of-the-art hardware architecture to support weight pruning \cite{yuan2018sticker,parashar2017scnn}. The hardware implementations are synthesized in SMIC 40nm CMOS process using Synopsys Design Compiler. Then we measure the delay values of those implementations. The speedup values of the pruning cases over the baseline are depicted in Fig. \ref{Fig:prune_speed}. In the figure, the speedup of the baseline itself is 1, and the results suggest that the pruning portion should be higher than about 55\%, in order to make sure that the benefits of pruning outperforms the overhead of indices. This corresponds to a break-even weight pruning ratio of 2.22.

\begin{figure}[htbp]
\centering
\includegraphics[width=0.45\textwidth]{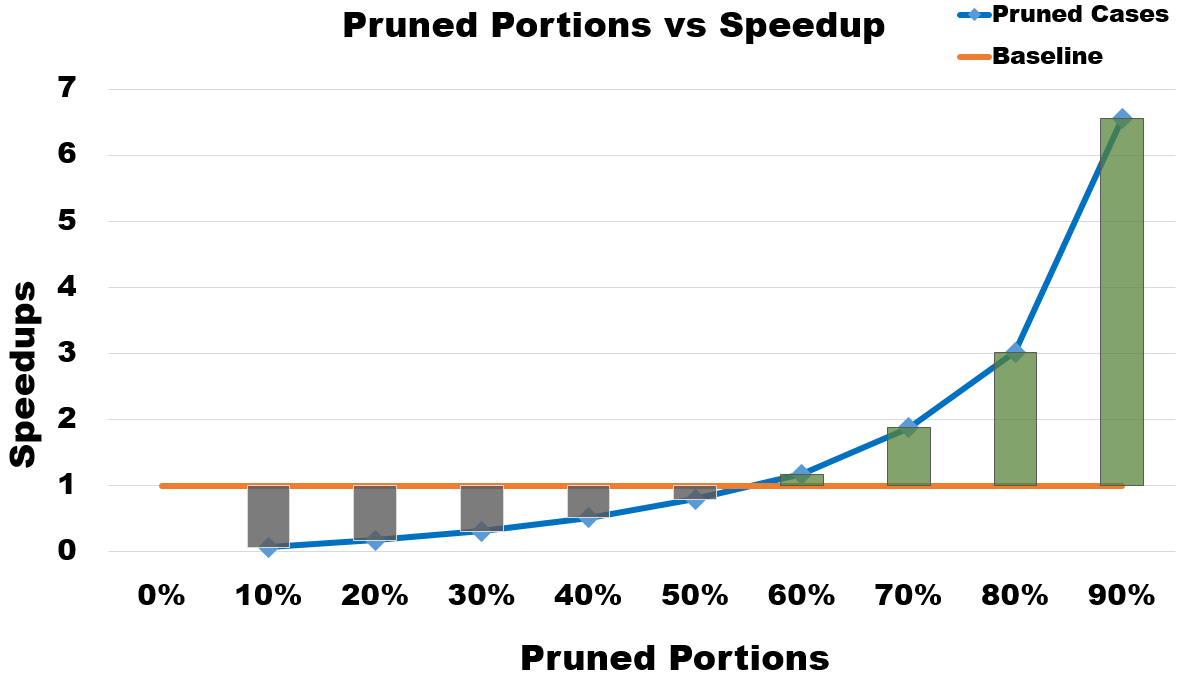}
\caption{Speedup comparison between pruned cases and baseline on a DNN layer, in order to derive the break-even weight pruning ratio.}
\label{Fig:prune_speed}
\end{figure}

\textbf{\emph{Hardware-Aware DNN Model Compression Algorithm:}} Based on the efficient calculation of such break-even pruning ratios, we develop efficient hardware-aware DNN model compression algorithm. We mainly focus on the CONV layers and perform weight pruning/quantization on FC layers accordingly to maintain accuracy. The detailed algorithm description is in Fig. \ref{Fig:AlgorithmHardware} as detailed in the following.

\begin{figure}[htbp]
\centering                                    
\includegraphics[width=0.51\textwidth]{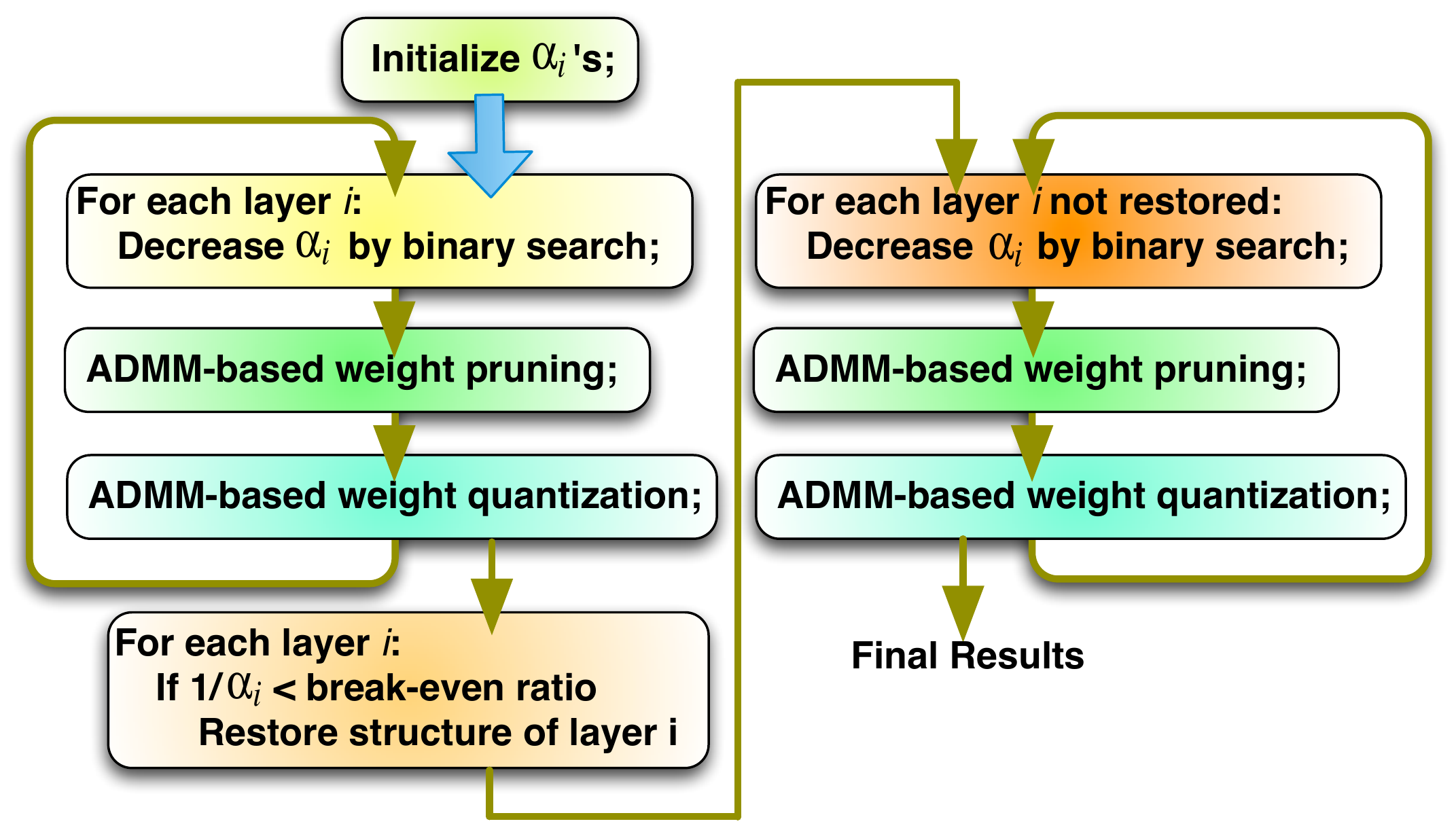} 
\caption{Algorithm of hardware-aware DNN model compression.}\label{Fig:AlgorithmHardware}
\vspace{-6mm}
\end{figure}

Consider a DNN with $N'$ CONV layers. Let $C_i$ ($1\le i\le N'$) denote the amount of computation, in the total number of operations, of the original DNN without weight pruning. Let $\alpha_i$ denote the portion of remaining weights in layer $i$ after weight pruning, and $\displaystyle \frac{1}{\alpha_i}$ denotes the pruning ratio in layer $i$. We start from pretrained DNN models, and initialize $\alpha_i$ values from those in the prior work such as \cite{han2015learning,han2015deep}, which can partially reveal the sensitivity to weight pruning for each layer. Since (i) our ADMM framework achieves higher performance and (ii) we focus mainly on CONV layers, we are able to reduce $\alpha_i$ values for different $i$. This is an iterative procedure. The amount of reduction $\Delta \alpha_i$ in each iteration is proportional to $C_i$. The underlying principle is to reduce the computation to a larger extent in those layers that are more computationally intensive (and likely, with a higher degree of redundancy). Binary search algorithm is exploited to find the updated $\alpha_i$ values that will not result in any accuracy degradation (this constraint can be relieved to a pre-defined accuracy degradation constraint). Please note that the FC layers will be pruned in accordance through this procedure for accuracy considerations.

The next step is to check whether the pruning ratios $\displaystyle \frac{1}{\alpha_i}$ surpass the hardware-specific break-even pruning ratio. If not then performing pruning on layer $i$ will not be beneficial for hardware acceleration. In this case we will (i) restore the structure for all layers that cannot surpass the break-even ratio (e.g., the first layer in AlexNet in practice), and (ii) reduce the $\alpha_i$ values of the other layers and perform ADMM-based weight pruning. Binary search is also utilized to accelerate the search. Upon convergence those layers will still surpass the break-even pruning ratio since we only decrease $\alpha_i$ values in the procedure.

After weight pruning, we perform ADMM-based weight quantization in order to further reduce computation and improve energy efficiency. Weight quantization is performed on both CONV and FC layers, but CONV layers will be given top priority in this procedure.

\section{Results and Discussions on Computation Reduction and Hardware-Aware Optimizations}

In this section, we first perform comparison on the computation reduction results focusing on the CONV layers (FC layers will be pruned accordingly as well to maintain accuracy). Next we compare on the synthesized hardware speedup results between the proposed hardware-aware DNN model compression algorithm with baselines. The baselines include the iterative weight pruning and weight clustering work \cite{han2015learning,han2015deep}, and recent work \cite{mao2017exploring,wen2016learning} of DNN weight pruning focusing on computation reductions. Due to space limitation, we only illustrate the comparison results on AlexNet (BVLC and CaffeNet models) on ImageNet dataset, but we achieve similar results on other benchmarks.
Again algorithm implementations are on the open-source Caffe tool with code/model release, and DNN model training and model compression are performed using NVIDIA 1080Ti and P100 GPUs.
\vspace{-3mm}

\begin{table*}[h]
\centering
\caption{Comparison results on the computation reduction with two metrics for the five CONV layers of AlexNet model.}\label{Table:AlexNetCONV}
\begin{tabular}{p{1.5cm}p{1.cm}p{1.cm}p{1.cm}p{1cm}p{1cm}p{1.5cm}p{1cm}p{1.cm}p{1.cm}p{2cm}}
\hline
 & & & \bf{MAC} & \bf{Operations} \\
\hline
 & CONV1 & CONV2 & CONV3 & CONV4 & CONV5 & CONV1-5 & FC1 & FC2 & FC3 & Overall prune \\ 
\hline
AlexNet & 211M & 448M & 299M & 224M & 150M & 1,332M & 75M & 34M & 8M & - \\
Ours & 133M & 31M & 18M & 16M & 11M & 209M & 7M & 3M & 2M & 13$\times$\\
Han \cite{han2015learning} & 177M & 170M & 105M & 83M & 56M & 591M & 7M & 3M & 2M & 9$\times$ \\
Mao \cite{mao2017exploring} & 175M & 116M & 67M & 52M & 35M & 445M & 5M & 2M & 1.5M & 12$\times$\\
Wen \cite{wen2016learning} & 180M &107M & 44M & 42M & 36M & 409M & 75M & 34M & 8M & 1.03$\times$\\
\hline
\hline
 & & & \bf{MAC} & \bf{$\times$} & \bf{bits}\\
\hline
Ours & 931M & 155M & 90M & 80M & 55M & 1,311M & - & - & - & -\\
Han \cite{han2015learning} & 1,416M & 1,360M & 840M & 664M & 448M & 4,728M & - & -& -& - \\
\hline
\end{tabular}
\end{table*}

\begin{table*}[h]
\centering
\caption{The synthesized hardware speedup for the five CONV layers of AlexNet model}\label{Table:AlexNetSpeedup}
\begin{tabular}{p{1.5cm}p{1.cm}p{1.cm}p{1.cm}p{1.cm}p{1.cm}p{2.8cm}p{2.8cm}p{2.5cm}}
\hline
 & CONV1 & CONV2 & CONV3 & CONV4 & CONV5 & CONV1-5 speedup & Conv1-5 prune ratio & Accuracy Degra. \\ 
\hline
AlexNet & 1$\times$ & 1$\times$ & 1$\times$ & 1$\times$ & 1$\times$ & 1$\times$ & 1$\times$ & 0.0\% \\
\hline
Ours1 & 1$\times$ & 7$\times$ & 7.5$\times$ & 7.2$\times$ & 7.1$\times$ & 3.6$\times$ & 13.1$\times$ & 0.0\%\\
\hline
Ours2 & 1$\times$ & 8.6$\times$ & 9.0$\times$ & 8.8$\times$ & 8.6$\times$ & 3.9$\times$ & 25.5$\times$ & 1.5\%\\
\hline
Han \cite{han2015learning} &  0.16$\times$ & 1.4$\times$ & 1.6$\times$ & 1.5$\times$ & 1.5$\times$ & 0.64$\times$ &2.7$\times$ & 0.0\%\\
\hline
Mao \cite{mao2017exploring} &  0.17$\times$ & 2.6$\times$ & 3$\times$ & 3$\times$ & 3$\times$ & 0.81$\times$ & 4.1$\times$& 0.0\%\\
\hline
Wen \cite{wen2016learning} &  0.15$\times$ & 2.9$\times$ & 4.6$\times$ & 3.8$\times$ & 2.9$\times$ & 0.77$\times$ & 5$\times$& 0.0\%\\
\hline
\end{tabular}
\end{table*}

\subsection{Computation Reduction Comparisons}

Table \ref{Table:AlexNetCONV} illustrates the comparison results on the computation reduction for the five CONV layers of AlexNet model. We show both layer-wise results and the overall results for all CONV layers. We use two metrics to quantify computation reduction. The first metric is the number of multiply-and-accumulation (MAC) operations, the key operations in the DNN inference procedure. This metric is directly related to the hardware performance (speed). The second metric is the product of the number of MAC operations and bit quantization width for each weight. This metric is directly related to the energy efficiency of (FPGA or ASIC) hardware implementation.

As can be observed in the table, the proposed ADMM framework achieves significant amount of computation reduction compared with prior work, even when some \cite{mao2017exploring,wen2016learning} also focus on computation reductions. For the first metric of computation reduction, the improvement can be close to 3$\times$ compared with prior work for CONV layers, and this improvement reaches 3.6$\times$ for the second metric. The improvement on the second metric of computation reduction is even higher because of the higher capability of the proposed method in weight quantization. We can also observe that the first CONV layer is more difficult for weight pruning and quantization compared with the other layers. This will impact the hardware speedup as shall be seen in the latter discussions. 

Because CONV layers are widely acknowledged to be more difficult to perform pruning than FC layers, the high performance in CONV layer pruning and quantization further demonstrates the effectiveness of the ADMM-based DNN model compression technique. Besides, although our results focus on CONV layer compression, we achieve 13$\times$ weight pruning ratio on the overall DNN model because FC layers are pruned as well. The overall weight pruning on DNN model is also higher than the prior work. The layer-wise pruning results are shown in Table \ref{Table:AlexNetCONV}. In this way we simultaneously achieve computation and model size reduction. 

\subsection{Synthesized Hardware Speedup Comparisons}

Table \ref{Table:AlexNetSpeedup} illustrates the comparison results, between the hardware-aware DNN model compression algorithm and baselines, on the synthesized hardware speedup for the five CONV layers of AlexNet model. The overall weight pruning ratio on the five CONV layers is also provided. We show both layer-wise results and the overall results for all CONV layers. The overall result is a weighted sum of the layer-wise results because of different amount of computation/parameters for each layer. The synthesized results are based on (i) the PE synthesis based on SMIC 40nm CMOS process using Synopsys Design Compiler, and (ii) the execution on a representative CONV layer (CONV4 of AlexNet). The hardware synthesis process accounts for the hardware performance overhead of weight pruning. Although the synthesis is based on ASIC setup, the conclusion generally holds for FPGA as well. For hardware speedup synthesis, we use the same number of PEs for the proposed method and baselines, and do not account for the advantage of the proposed method in weight quantization. This metric is conservative for the proposed method, but could effectively illustrate the effect of hardware-aware DNN optimization and the break-even pruning ratio.

In terms of hardware synthesis results, our methods result in speedup compared with original DNNs without compression. On the other hand, the baselines suffer from speed degradations. Such degradations are actually observed in prior work \cite{wen2016learning,han2015learning,yu2017scalpel}.
As can be observed from the table, we do not perform any weight pruning on the first CONV layer. This is because the weight pruning ratio for this layer is lower than the break-even pruning ratio derived in the previous section. In this way weight pruning will not bring about any speedup benefit for this layer. The underlying reason is that weights in the first CONV layer are directly connected to the pixels of the input image, and therefore most of the weights in the first CONV layer are useful. Hence the margin of weight pruning in the first CONV layer is limited. Although the first CONV layer is small compared with the other layers in terms of the number of weights, it will become the computation bottleneck among all CONV layers. This observation is also true in other DNNs like VGGNet or ResNet. When neglecting this factor, the baseline methods will incur degradation in the speed (which is common for all baselines in the first CONV layer) compared with the original DNN models without compression. Of course, speedups will be observed in baselines if they leave CONV1 unchanged.

When we target at further weight pruning on the CONV layers with certain degree of accuracy loss, we can achieve 25.5$\times$ weight pruning on overall CONV layers (40.5$\times$ pruning on CONV2-5) with only 1.5\% accuracy loss. In contrast to the significant pruning ratio, the synthesized speedup only has a marginal increase. This is because of the bottleneck of CONV1 and the saturation of speedup in the other CONV layers.

\section{Conclusion}

We present ADMM-NN, an algorithm-hardware co-optimization framework of DNNs using Alternating Direction Method of Multipliers (ADMM). 
The first part of ADMM-NN is a systematic, joint framework of DNN weight pruning and quantization using ADMM. 
The second part is hardware-aware optimizations to facilitate hardware-level implementations. We perform ADMM-based weight pruning and quantization accounting for (i) the computation reduction and energy efficiency improvement, and (ii) the performance overhead due to irregular sparsity. 
Exprimental results demonstrate that by combining weight pruning and quantization, the proposed framework can achieve 1,910$\times$ and 231$\times$ reductions in the overall model size on the LeNet-5 and AlexNet models. Highly promising results are also observed on VGGNet and ResNet models. Also, without any accuracy loss, we can achieve 3.6$\times$ reduction in the amount of computation, outperforming prior work.

\begin{acks}
This work is partly supported by the National Science Foundation (CNS-1739748, CNS-1704662, CCF-1733701, CCF-1750656, CNS-1717984, and CCF-1717754).
\end{acks}

\bibliographystyle{acm} 
\bibliography{template}

\end{document}